\documentclass[10pt,twocolumn,letterpaper]{article}

\usepackage[applications]{wacv}      

\definecolor{wacvblue}{rgb}{0.21,0.49,0.74}
\usepackage[pagebackref,breaklinks,colorlinks,allcolors=wacvblue]{hyperref}

\usepackage{graphicx}
\usepackage{amsmath}
\usepackage{amssymb}
\usepackage{booktabs}
\usepackage{multirow}
\usepackage{algorithm}
\usepackage{algorithmic}

\def\wacvPaperID{*****} 
\def\confName{WACV}
\def\confYear{2026}

\title{Unsupervised Incremental Learning Using Confidence-Based Pseudo-Labels}

\author{Lucas Rakotoarivony\\
Thales, cortAIx Labs\\
Palaiseau, 91120, France\\
{\tt\small lucas.rakotoarivony@thalesgroup.com}
}

\begin{document}
\maketitle

\begin{abstract}
Deep learning models have achieved state-of-the-art performance in many computer vision tasks. However, in real-world scenarios, novel classes that were unseen during training often emerge, requiring models to acquire new knowledge incrementally. Class-Incremental Learning (CIL) methods enable a model to learn novel classes while retaining knowledge of previous classes. However, these methods make the strong assumption that the incremental dataset is fully labeled, which is unrealistic in practice. In this work, we propose an unsupervised Incremental Learning method using Confidence-based Pseudo-labels (ICPL), which replaces human annotations with pseudo-labels, enabling incremental learning from unlabeled datasets. We integrate these pseudo-labels into various CIL methods with confidence-based selection and evaluate performance degradation on CIFAR100 and ImageNet100. Then, we compare our approach to popular Class Incremental Novel Category Discovery (class-iNCD) methods addressing similar challenges. Additionally, we apply our method to fine-grained datasets to demonstrate its real-world practicality and measure its computational complexity to validate its suitability for resource-constrained environments. ICPL achieves competitive results compared to supervised methods and outperforms state-of-the-art class-iNCD methods by more than 5\% in final accuracy.
\end{abstract}

\section{Introduction}
\label{sec:intro}
\begin{figure}[t]
	\centering
	\includegraphics[width=0.9\linewidth]{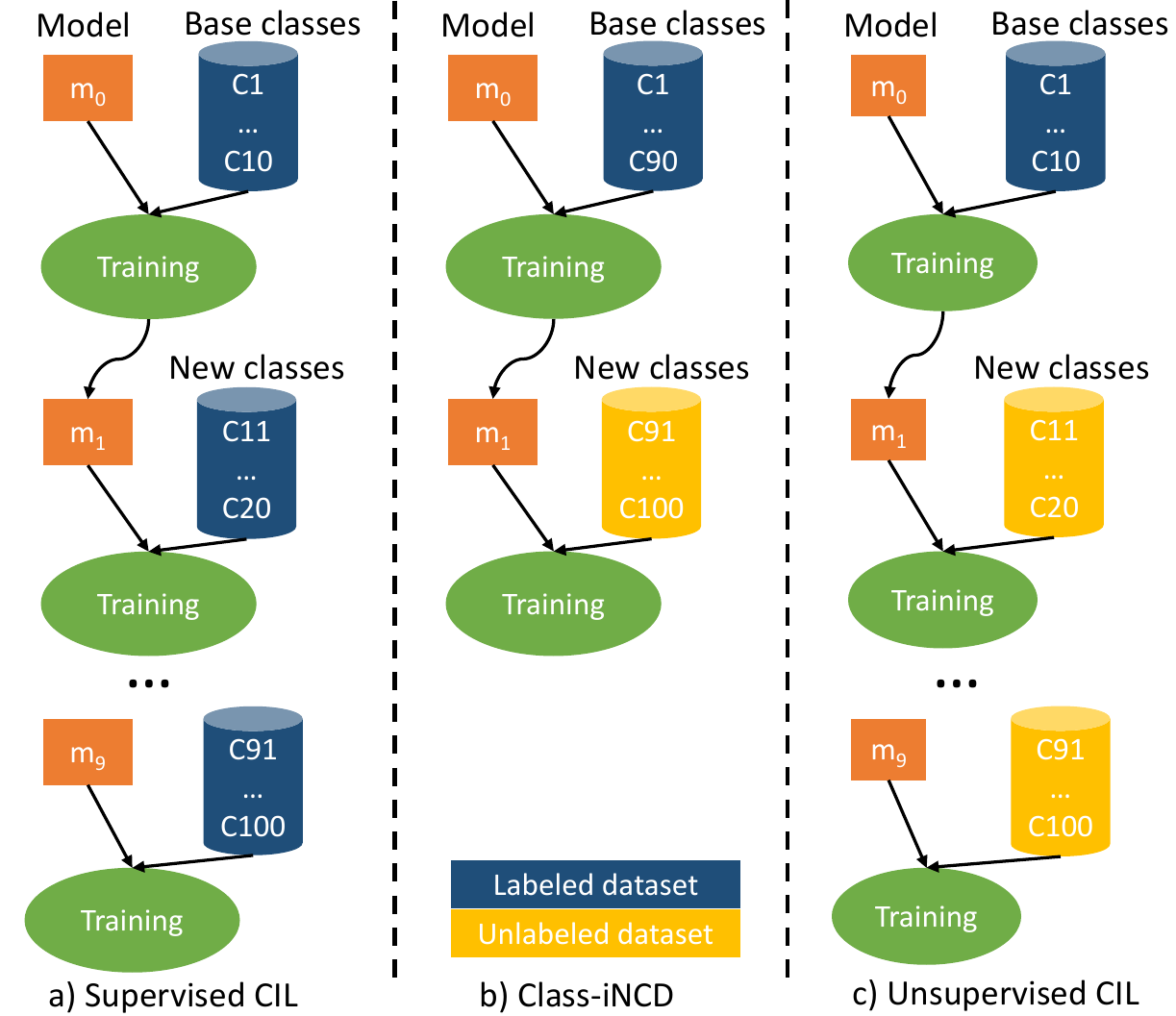} 
	\caption{Comparison of (a) supervised CIL, which uses a labeled dataset for each incremental task; (b) class-iNCD, which uses raw data during the incremental step but does not support many incremental tasks; and (c) our proposed unsupervised CIL, where no annotations are available for the new classes and multiple incremental tasks are considered.}
	\label{diff}
\end{figure}

Over the past few years, deep neural networks have demonstrated impressive performance across various applications, even surpassing human-level performance in most tasks. However, the environments used for training deep learning models are often much simpler than those encountered in real-world applications. For example, in a realistic environment, the number of classes is not fixed. However, neural networks can only recognize classes that were present during training, making it impossible for them to identify and classify new classes that may emerge in real-world streaming data. Class-Incremental Learning (CIL) methods \cite{li2017learning, rebuffi2017icarl, zhou2023deep} address this limitation by enabling models to learn new classes while retaining knowledge of previously learned ones. However, CIL methods rely on having a labeled dataset for each incremental update, a requirement that demands extensive human annotation, making the process time-consuming, labor-intensive, and costly. Class incremental Novel Category Discovery (class-iNCD) methods \cite{roy2022class, zhang2022grow}, attempt to bypass this need for human annotation by learning from unlabeled datasets. Yet, these approaches are not well-suited for handling a large number of incremental tasks. Unsupervised incremental learning \cite{khare2021unsupervised, he2021unsupervised}, which aims to enable models to learn new classes from raw images while preserving prior knowledge, is still a largely underexplored area. This task is challenging because it requires neural networks to address both catastrophical forgetting of prior knowledge when learning new classes and the presence of noisy labels, given the lack of human annotations. Figure \ref{diff} illustrates the difference between these three families of methods.

Therefore, in this work, we propose an unsupervised Incremental learning method using Confidence-based Pseudo-Labels, which we name ICPL, to address both issues. ICPL employs KMeans \cite{lloyd1982least} clustering algorithm on the embeddings of images from new classes to generate pseudo-labels. These pseudo-labels can then be integrated into popular CIL methods, eliminating the need for human annotations. 
While pseudo-labeling \cite{lee2013pseudo} is a common technique in unsupervised methods, it is often noisy due to potential errors. To mitigate this, we introduce a confidence selection method to use only high-confidence pseudo-labels. Inspired by \cite{karpusha2020calibrated, sikar2024accept}, we define confidence by applying a softmax transformation to the inverse of the KMeans matrix distance, providing a reliable measure of pseudo-label quality. 

Then, we combine ICPL with popular supervised CIL approaches, such as Replay \cite{zhou2023deep}, iCaRL \cite{rebuffi2017icarl}, WA \cite{zhao2020maintaining}, and FOSTER \cite{wang2022foster} to enable unsupervised incremental learning. It's important to note that we are not introducing new strategies to tackle catastrophic forgetting; our focus is on implementing an unsupervised approach.

Our experimental results show that, despite its simplicity, ICPL causes only a slight degradation in performance compared to the supervised case and outperforms class-iNCD strategies in long-term incremental learning.Additionally, we evaluate our method with fine-grained datasets to represent realistic use-cases and demonstrate that ICPL is effective across a wide variety of scenarios. Finally, we estimate the computational complexity of our method and show that ICPL reduces both the time and TFLOPs of the complete incremental training, making it suitable for resource-constrained environments.
Specifically, this method enables the model to learn multiple new classes without requiring human annotations, making it more practical for real-world applications.

The primary contributions of this paper are as follows:

\begin{enumerate}
	\item We develop a simple yet effective method, ICPL, for unsupervised Incremental learning that uses Confidence-based Pseudo-Labels instead of human annotations, making it relevant and challenging for real-world scenarios.
	\item Our ICPL method is implemented in several popular incremental learning methods, and we analyze the performance degradation compared to supervised methods to evaluate the effectiveness of our approach.
	\item Our results on CIFAR100 \cite{krizhevsky2009learning} demonstrate superior performance compared to state-of-the-art class-iNCD methods in long-term incremental learning setups.
\end{enumerate}

\section{Related work}

\subsection{Incremental learning}
\label{subsec:incremental}

Incremental learning is a challenging task where new instances are introduced during different training phases. However, the introduction and learning of new classes tend to suppress the knowledge acquired about the old classes. This problem, known as catastrophic forgetting \cite{mccloskey1989catastrophic}, is the main challenge for CIL methods. Various methods have been proposed to address this problem and are generally grouped into the following categories:

\textbf{Data centric}: These methods aim to retain the performance of previously learned classes by using a rehearsal dataset of exemplars from previous instances, in addition to the new dataset. The selection of exemplars is crucial, leading to various strategies, \cite{riemer2018learning, wu2019large} propose random selection, while \cite{bang2021rainbow} suggest selecting exemplars based on uncertainty computed with data augmentation. \cite{rebuffi2017icarl} use a herding strategy to choose exemplars that are closest to the mean feature representation of each class, while \cite{chaudhry2018riemannian} propose the opposite approach of selecting exemplars with high entropy near the decision boundary. 
Data replay methods store a much smaller exemplar set than the new class dataset, leading to a bias towards new instances. Therefore, data replay is often combined with the following methods.  

\textbf{Algorithm centric}: Knowledge distillation \cite{hinton2015distilling}, a popular solution to catastrophic forgetting in CIL, involves the old model distilling its features or logits into the current model to preserve knowledge of previous classes while learning new ones. Introduced by \cite{li2017learning}, this approach has since been refined with loss functions like KL Divergence for iCaRL \cite{rebuffi2017icarl} and cosine similarity-based loss for UCIR \cite{hou2019learning}.
Another approach to reduce bias towards new instances is modifying the weights associated to the new classes. UCIR \cite{hou2019learning} uses a cosine classifier to address the higher norm of new class weights, while WA \cite{zhao2020maintaining} equalizes weight norms between new and old classes.

\textbf{Model centric}: These methods allocate new parameters for each incremental task, thereby adjusting the model's capacity in order to fit the evolving data stream. DER \cite{yan2021dynamically} adds new feature extractors for each task and uses pruning to control model size. FOSTER \cite{wang2022foster} uses feature boosting and compression to increase capacity while avoiding feature redundancy. DyTox \cite{douillard2022dytox} achieves state-of-the-art performance with Transformers and dynamic task token expansion.

\subsection{Novel Category Discovery}

Novel Category Discovery (NCD) is a relatively recent task where a model must identify new classes within unlabeled data. Traditional NCD methods \cite{han2020automatically, han2021autonovel, zhao2021novel} focus solely on discovering new classes. In contrast, Generalized Class Discovery \cite{vaze2022generalized, fei2022xcon} assumes that the unlabeled dataset contains both new and old classes, requiring the model to recognize objects from previously seen categories while also identifying novel classes. However, these methods do not emphasize learning new classes, a challenge that class-iNCD methods address. For example, FRoST \cite{roy2022class} uses separate heads to learn from new data, while GM \cite{zhang2022grow} operates in a two-phase setup: a growing phase that increases feature diversity, followed by a merging phase to ensure satisfactory performance for known classes.

\section{Problem definition}
\label{sec:definition}

CIL is a complex task where new instances are introduced continuously in different training phases. We assume having a sequence of N training tasks $\{\mathcal{D}^1, \mathcal{D}^2, \ldots, \mathcal{D}^N\}$, where $\mathcal{D}^k\! =\! \{(x_i^k, y_i^k)\}_{i=1}^{n}$ represents the training dataset at step $k$ with $n$ training instances. Here, $x_i^k$ represents the i-th image of this task, and $y_i^k$ is the label within the label set $\mathcal{Y}^k$. We consider the classic CIL setup where there is no class overlap, so $\mathcal{Y}^k \cap \mathcal{Y}^{k'}\! =\! \emptyset$ for $ k \neq k'$. 

The majority of CIL methods use a rehearsal dataset $\mathcal{R}^k = \{(x_j^k, y_j^k)\}_{i=1}^{m}$ of size $m$ with $y_j^k \in (\mathcal{Y}^1 \cup \ldots \cup \mathcal{Y}^{(k-1)})$ in addition to the incremental dataset $\mathcal{D}^k$. This way, at each step $k$ the model can use $\mathcal{D}^k \cup \mathcal{R}^k$ for training. After each incremental training, the model is evaluated over all the previously seen classes $Y^k\! =\! (\mathcal{Y}^1 \cup \ldots \cup \mathcal{Y}^{k})$.

In contrast to supervised CIL, in our unsupervised case, we assume that for step $k\! \geq\! 2$, the labels are not available, so $\mathcal{D}^k \! =\! \{(x_i^k)\}_{i=1}^{n}$, while step 1 is used to initialize the network as in the supervised case. In the following sections, the pseudo-labels are denoted as $\hat{y}$.

As introduced in \cite{zhou2023deep}, we reuse the Base-m, Inc-n settings, where m stands for the number of classes learned in the first stage and n stands for the number of classes added in each incremental task. n is assumed to be known in our unsupervised setting, while label class information is not provided during training.

\section{Method description}
\subsection{Evaluation protocol through static encoding}
\label{subsec:evaluation}

\begin{figure*}[t]
	\centering
	\includegraphics[width=0.9\textwidth]{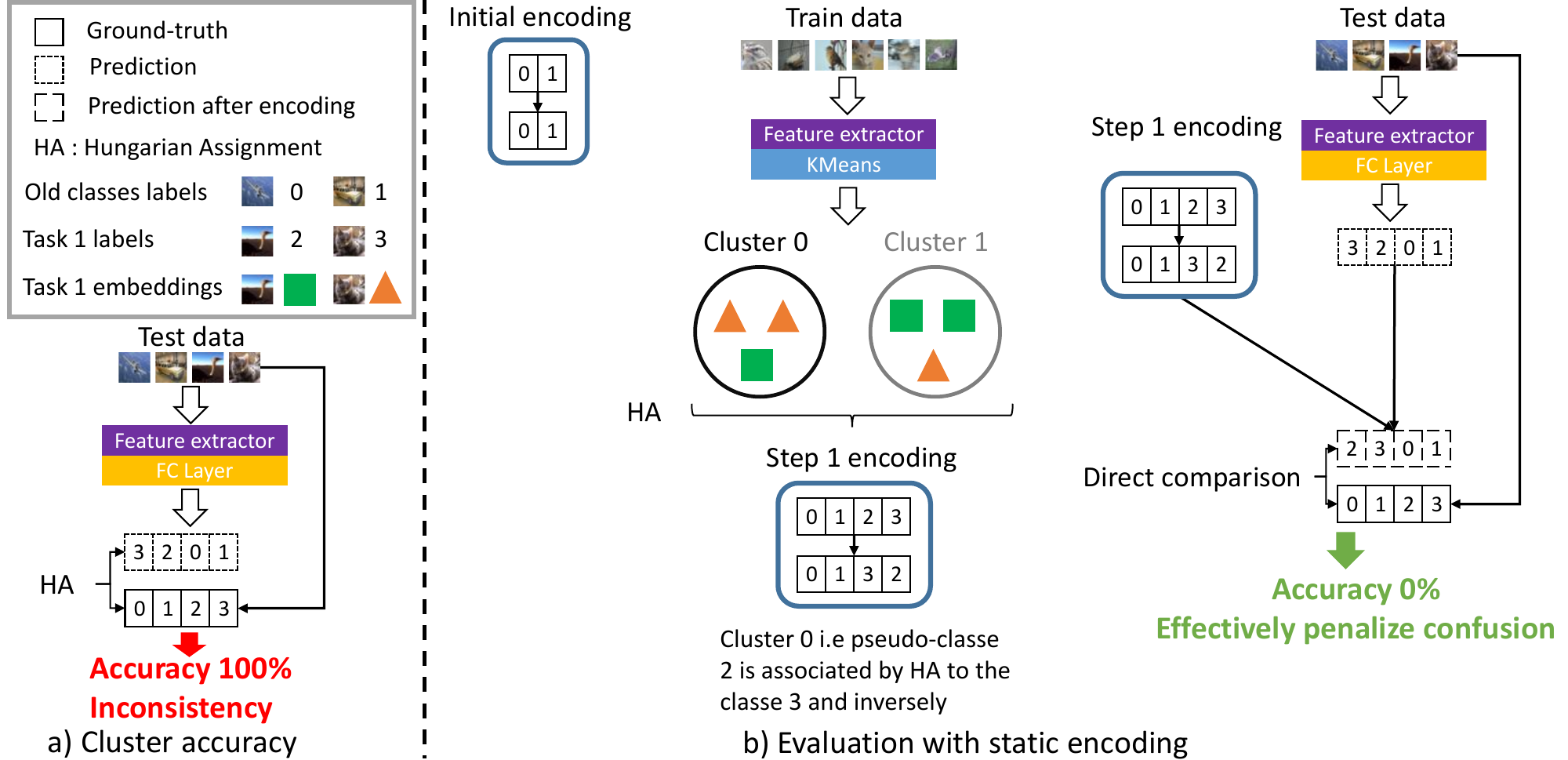}
	\caption{Comparison of evaluation protocol (a) Cluster accuracy evaluation, which obtains 100\% of accuracy while confusing old and new classes and (b) our proposed static encoding evaluation. During pseudo-label generation, a static encoding is obtained by applying the Hungarian assignment to match clusters with real classes. This encoding doesn't allow confusion between classes from different incremental tasks and ensures consistency during testing.}
	\label{metric}
\end{figure*}

Unsupervised incremental learning \cite{he2021unsupervised} and class-iNCD \cite{zhang2022grow} methods, generally use cluster accuracy based on Hungarian assignment \cite{kuhn1955hungarian} as a performance metric and consider the problem as a clustering task. However, this evaluation is inadequate in an unsupervised incremental setup because it allows confusion between instances from old and new classes. Indeed, Figure \ref{metric}(a) shows that the model classifies new class objects as old classes and inversely yet still achieves 100 \% of accuracy, which is unreasonable.   

FRoST \cite{roy2022class} has identified this problem and attempts to resolve it by using an encoding between the pseudo-class and the real class. However, their encoding is not static and can be modified at each incremental step, which is not ideal. Indeed, their method can modify at the second incremental step the encoding associated to the first incremental step even though no new data corresponding to the first step classes has been introduced. This is contrary to the idea that each pseudo-class corresponds to a ground-truth class.
  
In this work, we propose an evaluation protocol based on a static encoding. As shown in Figure \ref{metric}(b), we use the Hungarian assignment \cite{kuhn1955hungarian} to define an encoding between the clusters of generated pseudo-labels and the real classes. This encoding is fixed to prevent inconsistencies and is only updated by adding new associations when new classes are introduced. During testing, we apply this encoding to the predictions and directly compare the obtained results with the ground truth. Figure \ref{metric} demonstrates that our evaluation protocol is more coherent than previous protocols \cite{he2021unsupervised, roy2022class} and penalizes the metric when there is misclassification between classes of different tasks, which is the ideal behavior. We emphasize that this encoding requires labels but is only used for testing purposes, and therefore no label is required during unsupervised incremental training. Experimental results comparing cluster accuracy with our static encoding evaluation protocol are presented in \Cref{app:subsec:cluster}.

\begin{figure}[t]
	\centering
	\includegraphics[width=0.9\linewidth]{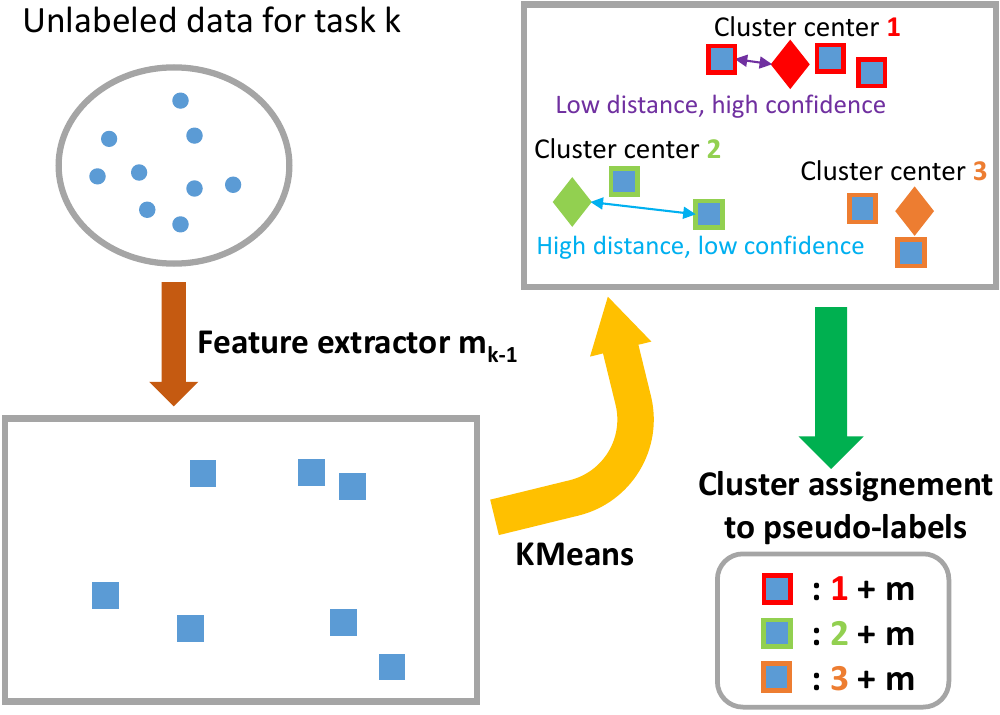}
	\caption{Illustration of the generation of pseudo-labels. Firstly, the feature extractor of task $k-1$ is used to generate embeddings from the unlabeled data. Then, KMeans algorithm generates clusters representing classes. Finally, we can use the distance to the cluster center to generate a confidence score for each pseudo-label.}
	\label{pseudo}
\end{figure}

\subsection{Generating pseudo-labels}

Clustering algorithms \cite{caron2018deep, xu2005survey} are a widely-used method for unsupervised learning due to their ability to extract valuable information from data with minimal supervision. In this work, we use KMeans \cite{lloyd1982least}, which is one of the most popular clustering methods and only requires the number of clusters. Clustering feature embeddings and its use in generating pseudo-labels have been proposed in various works \cite{caron2018deep, zhan2020online}. However, these works assume that the generated pseudo-labels correspond to already known classes. In contrast, we consider a dynamic environment where new classes can be introduced during training.

\begin{algorithm}[b]
	\caption{Confidence based Pseudo labels generation}
	\label{alg:algorithm}
    \begin{algorithmic}[1] 
		\REQUIRE Unlabeled Image set $\mathcal{D}^k\! =\! \{(x_i^k)\}_{i=1}^{n}$
		\REQUIRE Feature extractor $m_k$
		\REQUIRE Clustering algorithm KMeans $\Phi$
		\REQUIRE Confidence threshold $\alpha$
		\STATE Compute distance matrix  $D_{mat}\! =\! \Phi(m_k(\mathcal{D}^k)) $
		\STATE Get pseudo labels $\hat{\mathbf{Y}}\! =\! \{(\hat{y}_i^k)\}_{i=1}^{n}\! =\! \arg \min{D_{mat}}$
		\STATE Get standard deviation $\sigma\! =\! std(D_{mat})$
		\STATE Generate confidence values $ \mathbf{C}\! =\! \{(c_i^k)\}_{i=1}^{n}\! =\! \max_j \frac{\exp(-\frac{D_{mat_{ij}}^2}{2 \sigma^2})}{\sum_j \exp(-\frac{D_{mat_{ij}}^2}{2 \sigma^2})}$
		\STATE Select samples based on confidence $\hat{\mathcal{D}}^k\! =\! \{(x_i^k, \hat{y}_i^k) \mid c_i \geq \alpha \}_{i=1}^{n}$
		\STATE Return pseudo-labeled dataset $\hat{\mathcal{D}}^k\! =\! \{(x_i^k, \hat{y}_i^k)\}_{i=1}^{n}$
	\end{algorithmic}
	\label{pseudalg}
\end{algorithm}

Our ICPL framework proposes the following approach to generate pseudo-labels at step $k$ (with $k \geq 2$). 
Firstly, we apply the feature extractor $m_k$ obtained after incremental step $k\! -\! 1$ to the unlabeled data of the new incremental task to gather feature embeddings. Then, we use KMeans \cite{lloyd1982least} on these embeddings to generate clusters, with the idea that images of the same class should be relatively close in the embedding space \cite{cui2018survey, caron2018deep} thanks to their shared characteristics. The pseudo-labels are then derived based on the cluster assignment of their feature embeddings. 

It is commonly known ~\cite{rosenberg2005semi} that using all generated pseudo-labels during training can lead to a degradation in performance due to potential error in the pseudo-labels. 
From the $k$ clusters centers provided by KMeans and the previous embeddings, we generate a distance matrix $D_{mat}$ of size $(n,k)$ where $D_{mat}(i, j)$ represents the distance between the $i$-th cluster center and the $j$-th embedding. Inspired by previous work in deep metric learning \cite{karpusha2020calibrated, sikar2024accept}, we propose calculating a confidence score for each sample using a variant of the softmax function applied on the inverse of the generated distance matrix $D_{mat}$. Finally, we employ a confidence threshold $\alpha$ to select only the exemplars with high confidence for inclusion in the incremental training. Algorithm \ref{pseudalg} provides the pseudo-code for generating pseudo-labels, and Figure \ref{pseudo} presents a complete overview of the pseudo-label generation process.

\subsection{Incremental Learning with  pseudo-labels}
\label{subsec:integration}

As discussed in Section \ref{subsec:incremental}, CIL algorithms can be categorized into several types. We incorporate ICPL into the following representative algorithms from each category. Replay \cite{zhou2023deep} method (data centric) aim to reduce catastrophic forgetting by using a rehearsal dataset composed of samples from old classes. iCaRL \cite{rebuffi2017icarl} method (algorithm centric with knowledge distillation) uses feature distillation between the old and the new model to limit forgetting. WA \cite{zhao2020maintaining} (algorithm centric with model rectification) corrects bias towards new classes at each step by normalizing the weights of the fully connected layer. FOSTER \cite{wang2022foster} (model centric) uses feature boosting and compression to increase the model's capacity while avoiding parameter redundancy. These methods are popular incremental learning techniques that effectively represent various state-of-the-art strategies for mitigating catastrophic forgetting.

The incorporation of pseudo-labels into these methods is rather straightforward, it is achieved by substituting the $y$ label with the pseudo-label $\hat{y}$. The resulting $\hat{\mathcal{D}}^k\! =\! \{(x_i^k, \hat{y}_i^k)\}_{i=1}^{n}$ dataset is used to train the model on this incremental task. As proposed in \cite{zhang2021flexmatch}, we recalculate the pseudo-labels every $\tau$ epoch of incremental training and use data augmentation techniques like AutoAugment \cite{cubuk2018autoaugment} to enhance the feature extractor's discrimination ability. We use weight balancing \cite{xu2020class} and aggressive data augmentation like MixUp \cite{zhang2017mixup} for Replay and iCaRL, as we experimentally found that these methods can improve performance. However, these methods cause degradation with WA and FOSTER. A complete analysis of these phenomena is provided in Appendix \ref{app:ablation}, as well as an overview of the method.

\section{Experimental results}
\subsection{Experimental settings}

\textbf{Benchmark datasets} To remain consistent with prior studies, we evaluate ICPL on both the small-scale dataset CIFAR100 \cite{krizhevsky2009learning} and the large-scale dataset ImageNet-100 \cite{russakovsky2015imagenet}. However, recent work \cite{kim2023proxy} suggests using fine-grained datasets to better represent real-world scenarios. Therefore, we include additional experiments on such fine-grained datasets in \Cref{app:subsec:pretraining}. As mentioned in Section \ref{sec:definition}, we use the Base-m Inc-n training protocol proposed in \cite{zhou2023deep} to define the incremental configuration. Following the protocol defined in \cite{rebuffi2017icarl}, all classes are first shuffled using Numpy with a random seed 1993. For a fair comparison, we use the same training split and test split for every method.

\textbf{Implementation details} We implemented our ICPL method based on the code of \cite{zhou2023deep}, using PyTorch \cite{paszke2019pytorch} and Scikit-Learn \cite{pedregosa2011scikit} for KMeans. As proposed in \cite{rebuffi2017icarl}, we used a custom version of ResNet32 \cite{he2016deep} for CIFAR100 and ResNet18 for ImageNet100. During our comparison with the class-iNCD and GCD methods, we used a ResNet18 backbone to ensure a fair comparison. Following CIL conventions, we did not use pretrained models. However, in \Cref{app:subsec:pretraining}, we present experimental results that examine the effect of pretraining on the proposed ICPL method. We used AutoAugment \cite{cubuk2018autoaugment} to enhance the model's discrimination capability and regenerated pseudo-labels every $\tau\! =\! 10$ epochs. The confidence threshold $\alpha$ for selecting pseudo-labels was set at 0.85. For the iCaRL and Replay methods, we used MixUp \cite{zhang2017mixup} and class weighting \cite{xu2020class} to mitigate catastrophic forgetting.
We used SGD with an initial learning rate of 0.1 and momentum of 0.9. The training epoch was set to 170 for all datasets, with a batch size of 128. The learning rate decayed by 0.1 at 80 and 120 epochs. The exemplar set size was set to 2000, as proposed by \cite{zhou2023deep}, for all datasets. These exemplars were equally sampled from each seen class via the herding \cite{welling2009herding} algorithm. All training was executed on a single Nvidia GeForce RTX 3090.

\textbf{Evaluation metrics} Previous works uses cluster accuracy for evaluation \cite{he2021unsupervised, zhang2022grow}; however, as discussed in Section \ref{subsec:evaluation}, this is not an adequate evaluation protocol for unsupervised incremental learning. With our proposed static encoding evaluation protocol (see Figure \ref{metric}), we can directly measure the Top-1 accuracy by comparing the encoded model predictions to the ground truth labels. We always use the prediction of the classifier and do not use the NME \cite{rebuffi2017icarl} classifier, even with the iCaRL method.
We define $\Theta_k$ as the Top-1 accuracy after the $k$-th incremental task. Following previous work \cite{zhou2023deep, he2021unsupervised}, we use the final accuracy $\Theta_N$ as a metric to show the model's ability to retain information, and the average accuracy $\tilde{\Theta}\! =\! \frac{1}{N}\sum_{i=1}^{N} \Theta_i$, which takes into account performance in each incremental phase.

\subsection{Comparison with supervised incremental learning}

\begin{table*}[]

	\centering
	\begin{tabular}{@{}ccccccccc|cccc@{}}
		\toprule
		\multirow{3}{*}{Method} & \multicolumn{8}{c}{CIFAR100}                                                                                                          & \multicolumn{4}{c}{ImageNet100}                                    \\
		& \multicolumn{2}{c}{Base0 Inc5} & \multicolumn{2}{c}{Base0 Inc10} & \multicolumn{2}{c}{Base0 Inc20} & \multicolumn{2}{c}{Base50 Inc10} & \multicolumn{2}{c}{B0 Inc10} & \multicolumn{2}{c}{B50 Inc10} \\ \cmidrule(l){2-3} \cmidrule(l){4-5} \cmidrule(l){6-7} \cmidrule(l){8-9} \cmidrule(l){10-11} \cmidrule(l){12-13}
		& $\Theta_N$           & $\tilde{\Theta}$          & $\Theta_N$           & $\tilde{\Theta}$           & $\Theta_N$           & $\tilde{\Theta}$           & $\Theta_N$            & $\tilde{\Theta}$          & $\Theta_N$          & $\tilde{\Theta}$          & $\Theta_N$            & $\tilde{\Theta}$          \\ \midrule \midrule
		Replay \cite{zhou2023deep}                  & 35.69              & 55.14             & 40.07              & 58.46              & 42.71              & 59.84              & 42.05               & 51.84              & 40.2             & 58.60             & 39.38               & 51.77             \\
		Replay + ICPL           & \underline{30.02}              & 40.75             & \underline{31.57}              & 48.40              & 30.13              & 46.76              & 30.01               & 40.27              & \underline{34.34}             & \underline{52.33}             & \underline{34.3}               & \underline{45.76}             \\ \midrule
		iCaRL \cite{rebuffi2017icarl}                   & 36.99              & 56.13             & 41.09              & 59.34              & 46.35              & 61.85              & 44.5               & 55.26              & 41.14             & 59.89             & 41.24               & 55.29             \\
		iCaRL + ICPL            & \underline{30.79}             &  41.16            & \underline{32.03}              & 49.00              & 33.32              & 50.86              & \underline{37.3}               & \underline{48.15}              & \underline{32.92}             & 47.91             & \underline{35.9}               & \underline{48.49}             \\ \midrule
		WA \cite{zhao2020maintaining}                     & 45.67              & 61.91             & 51.38              & 64.79              & 54.6              & 66.17              & 53.8               & 62.96              & 53.02             & 68.06             & 55.08               & 65.78             \\
		WA + ICPL               & 33.78           & 51.87            & 33.5            & 48.65            & 36.78           & 53.22            & \underline{45.86}            & \underline{57.74}          & 38.54             & 54.83             & \underline{45.8}               & \underline{56.45}             \\ \midrule
		FOSTER \cite{wang2022foster}                  & 47.61              & 62.37             & 53.65              & 66.46              & 60.22              & 70.18              & 58.41               & 67.01              & 59.72             & 69.23             & 64.62               & 73.02             \\
		FOSTER + ICPL           & \underline{39.23}           & \underline{53.24}            & 35.57           & 50.00            & 40.9            & 55.64           & \underline{55.13}             & \underline{65.68}           & 44.06             & 58.55             & \underline{60.9}               & \underline{69.63}     \\ \bottomrule[0.5pt]     
	\end{tabular}
    \caption{Comparison of supervised and unsupervised incremental learning, with the unsupervised approach incorporating ICPL, across various CIL methods and different incremental settings. Degradation inferior to 10\% is underlined}
	\label{degradation}
\end{table*}

\begin{figure*}[]
	\centering
	\includegraphics[width=1.0\textwidth]{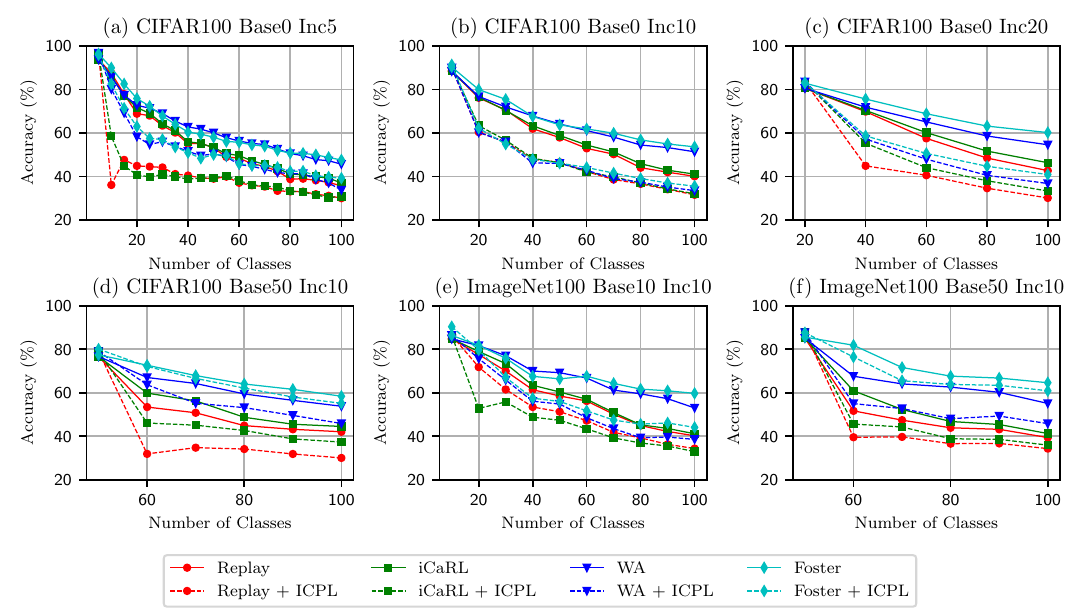}
	\caption{Incremental accuracy curves for supervised and unsupervised cases on CIFAR100 and ImageNet100 datasets.}
	\label{incremental}
\end{figure*}

In this section, we evaluate the effectiveness of our ICPL method compared to supervised approaches. Using the methodology detailed in Section \ref{subsec:integration}, we integrate ICPL with the CIL methods Replay \cite{zhou2023deep}, iCaRL \cite{rebuffi2017icarl}, WA \cite{zhao2020maintaining}, and FOSTER \cite{wang2022foster}. Table \ref{degradation} summarizes the results in terms of final accuracy $\Theta_N$ and average accuracy $\tilde{\Theta}$ for various incremental settings and datasets, while Figure \ref{incremental} presents the incremental accuracy curves. "Replay + ICPL" represents the Replay method combined with ICPL to perform unsupervised incremental learning. 

As expected, we observe a performance degradation compared to the supervised case, which can be explained by both the quality (errors in the pseudo-labels) and quantity (not using all pseudo-labels) in comparison to human annotations. However, this degradation remains relatively low, with an average of 10.27\% for the final accuracy and 10.25\% for the average accuracy, which is reasonable given the time and resource savings achieved by our method compared to manual annotation. 
Our method benefits from a larger number of base classes, as this allows the feature extractor to have learned more classes and features, resulting in better feature discrimination.
Using the Base50 Inc10 setup on the CIFAR100 dataset, FOSTER + ICPL achieves impressive results, with only a 1.33\% drop in average accuracy. It also surpasses supervised methods like WA under the same conditions, achieving 65.68\% of average accuracy compared to WA's 62.96\%.

ICPL is not limited to these CIL methods and shows competitive results compared to the supervised case while eliminating the need for fully human-annotated datasets at each incremental task, showing the interest of our method.

\subsection{Comparison with state-of-the-art class-iNCD methods}

\begin{table}[]
	\centering
    \small
	\begin{tabular}{@{}ccccccc@{}}
		\toprule
		\multirow{2}{*}{Method} & \multicolumn{2}{c}{Base0 Inc5} & \multicolumn{2}{c}{Base0 Inc10} & \multicolumn{2}{c}{Base50 Inc10} \\ \cmidrule(l){2-3} \cmidrule(l){4-5} \cmidrule(l){6-7}
		& $\Theta_N$           & $\tilde{\Theta}$          & $\Theta_N$           & $\tilde{\Theta}$         & $\Theta_N$           & $\tilde{\Theta}$                    \\ \midrule \midrule
		FRoST \cite{roy2022class}                  & 2.29              & 11.40             & 6.39              & 18.79              &  20.31               & 41.23              \\
		GM \cite{zhang2022grow}                     & 3.91              & 13.93             &   6.00            & 20.52            &  35.14               & 51.57  \\ \midrule 
        DCCL \cite{pu2023dynamic}                     & 9.78              & 20.62             &   12.61            & 33.20          &  40.85               & 54.03  \\
        CMS \cite{choi2024contrastive}                     & 12.89              & 25.37             &   18.63            & 37.22            &  41.18               & 54.47  \\ \midrule
		ICPL               & \textbf{38.53}              & \textbf{52.00}             & \textbf{35.99}              & \textbf{48.60}        & \textbf{46.81}               & \textbf{58.71}              \\ \bottomrule[0.5pt]
	\end{tabular}
    \caption{Comparison of ICPL with state-of the art class-iNCD and GCD methods.}
	\label{discovery}
\end{table}

Class-iNCD methods like FRoST \cite{roy2022class} and GM \cite{zhang2022grow}, and GCD methods such as DCCL \cite{pu2023dynamic} or CMS \cite{choi2024contrastive}, aim to discover new classes while considering Class-Incremental training. However, as illustrated in Figure \ref{diff}, these methods differ from our unsupervised incremental training setup because they do not consider long-term incremental learning. For fair comparison, we adapt these methods by using same ResNet18 backbone and our static encoding evaluation protocol as proposed in Section \ref{subsec:evaluation}.

Table \ref{discovery} compares ICPL with both class-iNCD and GCD methods for popular incremental learning settings. In the following experiments, ICPL refers to the combination of the WA method with our ICPL approach. In all setups, ICPL outperforms all methods by over 5\% for final accuracy, especially in cases with few base classes (Base0 Inc5 and Base0 Inc10), demonstrating that class-iNCD and GCD methods are not adequate for long-term incremental learning. In Base50 scenarios, GM, DCCL and CMS showed competitive but inferior results, particularly in maintaining high average accuracy, while FRoST failed to overcome catastrophic forgetting.

These results show that ICPL, without bells and whistles, effectively outperforms state-of-the-art class-iNCD and GCD methods and is able to maintain high accuracy across many incremental tasks.

\section{Ablation Study}

\subsection{Impact of pretraining}
\label{app:subsec:pretraining}

Using a pretrained ImageNet model \cite{he2019rethinking} is common in deep learning because it allows for better initialization and generalization. However, in incremental learning, it is common to train the model from scratch due to the potential overlap between ImageNet classes and novel classes. Recent work \cite{kim2023proxy} suggests using pretraining in incremental learning when dealing with fine-grained datasets due to the model's specialization towards highly specific features. Table \ref{fine} shows the results for three fine-grained datasets (Stanford Cars \cite{krause20133d}, FGVC-Aircraft \cite{maji13fine-grained} and Oxford-IIIT Pet \cite{parkhi2012cats}) when using ICPL with and without ImageNet pretraining, compared to the supervised incremental learning approach. Overall, ICPL demonstrates competitive performance compared to supervised approaches, with an average final accuracy reduction of just 9.75\%. However, when using ImageNet pretraining, we notice a slight degradation compared to the standard ICPL, which might seem counterintuitive. We argue that the model becomes too generalized after ImageNet pretraining, losing the ability to distinguish subtle differences crucial for classifying fine-grained datasets.

\begin{table}[]
	\centering
	\small
	\begin{tabular}{@{}ccc|cc|cc@{}}
		\toprule
		\multirow{3}{*}{Method}  & \multicolumn{2}{c}{Stanford Cars}   & \multicolumn{2}{c}{FGVC-Aircraft }    & \multicolumn{2}{c}{Oxford-IIIT Pet }    \\
		& \multicolumn{2}{c}{Base96 Inc10} & \multicolumn{2}{c}{Base50 Inc10} & \multicolumn{2}{c}{Base22 Inc5} \\ \cmidrule(l){2-3} \cmidrule(l){4-5} \cmidrule(l){6-7}
		& $\Theta_N$           & $\tilde{\Theta}$          & $\Theta_N$           & $\tilde{\Theta}$           & $\Theta_N$           & $\tilde{\Theta}$                   \\ \midrule \midrule
		Supervised               & 55.09               & 60.35                           & 58.78               & 61.63              & 55.86               & 58.87              \\ \midrule
		ICPL                     & \textbf{40.57}               & \textbf{50.34}              & \textbf{46.53}               & \textbf{55.08}              & \textbf{53.38}               & \textbf{54.18}              \\
		ICPL + PI & 38.29               & 49.22                      & 41.76               & 47.53              & 49.61               & 50.97             \\ \bottomrule[0.5pt]
	\end{tabular}
    \caption{Study of the impact of pretraining for fine-grained dataset. Average accuracy $\tilde{\Theta}$ and final accuracy $\Theta_N$ are reported. PI refers to Pretrain with ImageNet.}
	\label{fine}
\end{table}

\subsection{Effect of evaluation protocols}
\label{app:subsec:cluster}

As explained in Section \ref{subsec:evaluation}, and illustrated in Figure \ref{metric}, the current evaluation protocol based on cluster accuracy is not well suited for unsupervised incremental learning. Table \ref{tab:evaluation} shows the final and mean accuracy differences between the cluster accuracy-based and our proposed static encoding evaluation protocol. As expected, the cluster accuracy evaluation protocol consistently reports higher accuracy, but the differences are small (0.17\% to 2.10\% for $\Theta_N$ and 0.11\% to 1.30\% for $\tilde{\Theta}$), suggesting this issue may be optional. However, we believe that as unsupervised incremental learning and novel category discovery evolve toward more realistic use cases, class confusion will become more prevalent, especially in fine-grained datasets \cite{krause20133d, maji13fine-grained, parkhi2012cats}, making a more rigorous evaluation protocol necessary. 

The theoretical confusion in cluster accuracy, as discussed in \cite{roy2022class}, allows models to predict novel classes as old ones (and vice versa) without penalty, achieving 100\% accuracy. For example, in an autonomous vehicles context, cluster accuracy evaluation protocol would allow a model designed to recognize stop signs (old classes) to mistakenly identify a green light (new class) as a stop sign, as long as the reverse is true, without penalizing the model. This scenario, clearly undesirable, underscores the need for a more robust evaluation method.

\begin{table}[b]
\centering
\scriptsize
\begin{tabular}{@{}cccccccc@{}}
\toprule
\multirow{2}{*}{Method} & \multirow{2}{*}{Clustering} & \multicolumn{2}{c}{Base0 Inc5} & \multicolumn{2}{c}{Base0 Inc10} & \multicolumn{2}{c}{Base50 Inc10} \\ \cmidrule(l){3-4} \cmidrule(l){5-6} \cmidrule(l){7-8} 
		& & $\Theta_N$           & $\tilde{\Theta}$          & $\Theta_N$           & $\tilde{\Theta}$                   & $\Theta_N$            & $\tilde{\Theta}$                    \\ \midrule \midrule
\multirow{3}{*}{Replay} & Hierarchical & \textbf{33.46} & \textbf{51.47} & 30.07 & 45.32 & \textbf{37.82} & \textbf{50.54} \\
& Spectral     & 33.09 & 50.04 & 28.29 & 42.94 & 35.25 & 48.28\\
& KMeans       & 30.02 & 40.75& \textbf{31.57} & \textbf{48.40} & 30.01 & 40.27 \\ \midrule
\multirow{3}{*}{iCaRL} & Hierarchical  & \textbf{35.82} & \textbf{54.01} & \textbf{32.34} & 47.40 & \textbf{39.05} & \textbf{52.23} \\
& Spectral      & 31.94 & 49.62 & 28.06 & 42.91 & 37.61 & 51.69\\
& KMeans        & 30.79 & 41.16 & 32.03 & \textbf{49.00} & 37.30 & 48.15 \\ \midrule
\multirow{3}{*}{WA} & Hierarchical     & 34.51 & 50.74 & 33.49 & 48.43 & 44.08 & 57.69 \\
& Spectral         & \textbf{34.53} & 49.18 & 29.58 & 44.00 & 44.05 & 56.47 \\
& KMeans           & 33.78 & \textbf{51.87} & \textbf{33.50} & \textbf{48.65} & \textbf{45.86} & \textbf{57.74} \\ \midrule
\multirow{3}{*}{FOSTER} & Hierarchical & 38.09 & 51.25 & 35.56 & 48.88 & 53.94 & 65.46 \\
& Spectral     & 34.78 & 47.24 & 32.66 & 46.17 & 52.15 & 63.45 \\
& KMeans       & \textbf{39.23} & \textbf{53.24} & \textbf{35.57} & \textbf{50.00} & \textbf{55.13} & \textbf{65.68}\\
\bottomrule
\end{tabular}
\caption{Comparison of the average accuracy $\tilde{\Theta}$ and final accuracy $\Theta_N$ of ICPL on CIFAR100, using different clustering algorithms. In all configurations, the confidence threshold $\alpha$ is set at 0.85.}
\label{tab:clustering}
\end{table}

\begin{table*}[]
\centering
\begin{tabular}{@{}cccccccccc@{}}
\toprule
\multirow{2}{*}{Method} & \multirow{2}{*}{Evaluation protocol} & \multicolumn{2}{c}{Base0 Inc5} & \multicolumn{2}{c}{Base0 Inc10} & \multicolumn{2}{c}{Base0 Inc20} & \multicolumn{2}{c}{Base50 Inc10} \\ \cmidrule(l){3-4} \cmidrule(l){5-6} \cmidrule(l){7-8} \cmidrule(l){9-10}
		& & $\Theta_N$           & $\tilde{\Theta}$          & $\Theta_N$           & $\tilde{\Theta}$           & $\Theta_N$           & $\tilde{\Theta}$           & $\Theta_N$            & $\tilde{\Theta}$                    \\ \midrule \midrule
\multirow{2}{*}{Replay} & Cluster accuracy & 31.55 & 41.20 & 32.83 & 48.92 & 30.49 & 46.94 & 30.23 & 40.40 \\
& Static encoding       & \textit{30.02} & 40.75 & \textit{31.57} & 48.40 & 30.13 & 46.76 & 30.01 & 40.27 \\ \midrule
\multirow{2}{*}{iCaRL} & Cluster accuracy  & 32.00 & 41.71 & 32.45 & 49.27 & 34.66 & 52.16 & 37.75 & 48.87 \\
& Static encoding        & \textit{30.79} & 41.16 & 32.03 & 49.00 & \textit{33.32} & \textit{50.86} & 37.30 & 48.15 \\ \midrule
\multirow{2}{*}{WA} & Cluster accuracy     & 35.88 & 52.28 & 34.20 & 48.76 & 37.23 & 53.51 & 46.08 & 57.88 \\
& Static encoding           & \textit{33.78} & 51.87 & 33.50 & 48.65 & 36.78 & 53.22 & 45.86 & 57.74 \\ \midrule
\multirow{2}{*}{FOSTER} & Cluster accuracy & 39.49 & 53.38 & 36.40 & 50.33 & 41.07 & 55.78  & 55.42 & 65.85 \\
& Static encoding       & 39.23 & 53.24 & 35.57 & 50.00 & 40.90 & 55.64 & 55.13 & 65.68 \\
\bottomrule
\end{tabular}
\caption{Comparison of cluster accuracy and static encoding evaluation protocol using CIFAR100. Difference superior to 1\% is italicized.}
\label{tab:evaluation}
\end{table*}

\subsection{Comparison of clustering algorithms}

We selected KMeans as our clustering algorithm due to its simplicity and effectiveness. However, ICPL is flexible and can easily incorporate other clustering methods to generate pseudo-labels. As shown in the Table \ref{tab:clustering}, ICPL achieves comparable performance across different clustering techniques, including spectral clustering \cite{von2007tutorial} and hierarchical clustering \cite{mullner2011modern}. Although spectral clustering generally yields lower performance, hierarchical clustering often outperforms KMeans. Nonetheless, the differences are relatively minor, indicating that ICPL is robust to the choice of clustering algorithm and that this decision is not critical for achieving strong performance.

\subsection{Computational complexity analysis}

\begin{table}[t]
	\centering
    \small
	\begin{tabular}{@{}ccccc@{}}
		\toprule
		Method & TFLOPs $\downarrow $& Training time (s) $\downarrow$ & $\Theta_N \uparrow$          & $\tilde{\Theta} \uparrow$\\ \midrule \midrule
		Supervised                  & 4828            & 2060            &   51.38            & 
		64.79                                         \\ \midrule
		ICPL $\tau\! =\! \varnothing$            & \textbf{2668}              & \textbf{1525}             & 29.94             & 44.72   \\
		ICPL $\tau\! =\! 10$                     & 3672              & 2042             &   \textbf{33.50}            & \textbf{48.65}               \\
		ICPL $\tau\! =\! 20$                     & 3433              & 1948             &   31.66            & 47.30               \\
        	ICPL $\tau\! =\! 30$                     & 3730              & 2038             &   32.03            & 47.33 \\
		 \bottomrule                                                                             
	\end{tabular}
    \caption{Measurement of computational efficiency and training time with different step size $\tau$. $\tau\! =\! \varnothing$ means that pseudo-labels are only calculated once at the start of the incremental step. The models are trained with the WA method and a Base10 Inc10 setup.}
	\label{computation}
\end{table}

With the introduction of a new step in the incremental training, it is essential to analyze its impact on training time and computational complexity. Table \ref{computation} compares the supervised method with our unsupervised approach, using different step sizes for pseudo-label recomputation. Interestingly, the supervised approach exhibits the longest training time and highest TFLOPs in the training pipeline. While this may initially seem counterintuitive, as explained in Appendix \ref{app:subsec:gflop}, generating pseudo-labels is computationally efficient and incurs a relatively low cost compared to a full training step. Furthermore, since ICPL selects only high-confidence pseudo-labels and excludes certain data (approximately 30\% for $\alpha = 0.85$), the unsupervised method reduces both computational complexity and training time. For a step size of 10, it achieves a 24\% reduction in computational cost compared to the supervised approach, while also shortening the training time by a few seconds. 

We believe that training time could be further reduced by adopting a CUDA-accelerated KMeans implementation, as suggested by \cite{wu2024ft, li2023large}, instead of the Scikit-Learn \cite{pedregosa2011scikit} version. These findings demonstrate that ICPL is well-suited for practical applications, even in resource-constrained environments.

\section{Conclusion}

In this paper, we address the problem of unsupervised incremental learning, a scenario that is more realistic than the supervised case, as it eliminates the need for manual annotations during each incremental task. We proposed ICPL, a simple yet effective method that uses the KMeans algorithm to generate pseudo-labels for unlabeled data while introducing a fair evaluation protocol tailored to our problem.
We integrated ICPL with various CIL approaches and demonstrated that ICPL achieves competitive results compared to supervised methods and outperforms state-of-the-art class-iNCD and GCD approaches. Furthermore, we showed that ICPL is particularly effective in fine-grained scenarios, highlighting its relevance for real-world applications. Finally, its confidence-based data selection reduces both complexity and training time, making it suitable for resource-constrained environments.
Given the potential time and resource savings compared to the supervised approach, along with the promising results of ICPL, we hope this work will serve as a baseline for unsupervised incremental learning.

\section*{Acknowledgements}

This work is partially funded by the European Union, NEUROKIT2E (GrantNo.101112268).

{
    \small
    \bibliographystyle{ieeenat_fullname}
    \bibliography{icpl}

@String(ECCV= {Eur. Conf. Comput. Vis.})

@String(ECCV  = {ECCV})

@inproceedings{rebuffi2017icarl,
	title={icarl: Incremental classifier and representation learning},
	author={Rebuffi, Sylvestre-Alvise and Kolesnikov, Alexander and Sperl, Georg and Lampert, Christoph H},
	booktitle={Proceedings of the IEEE conference on Computer Vision and Pattern Recognition},
	pages={2001--2010},
	year={2017}
}

@article{riemer2018learning,
	title={Learning to learn without forgetting by maximizing transfer and minimizing interference},
	author={Riemer, Matthew and Cases, Ignacio and Ajemian, Robert and Liu, Miao and Rish, Irina and Tu, Yuhai and Tesauro, Gerald},
	journal={arXiv preprint arXiv:1810.11910},
	year={2018}
}

@inproceedings{wu2019large,
	title={Large scale incremental learning},
	author={Wu, Yue and Chen, Yinpeng and Wang, Lijuan and Ye, Yuancheng and Liu, Zicheng and Guo, Yandong and Fu, Yun},
	booktitle={Proceedings of the IEEE/CVF conference on computer vision and pattern recognition},
	pages={374--382},
	year={2019}
}

@inproceedings{bang2021rainbow,
	title={Rainbow memory: Continual learning with a memory of diverse samples},
	author={Bang, Jihwan and Kim, Heesu and Yoo, YoungJoon and Ha, Jung-Woo and Choi, Jonghyun},
	booktitle={Proceedings of the IEEE/CVF conference on computer vision and pattern recognition},
	pages={8218--8227},
	year={2021}
}

@inproceedings{chaudhry2018riemannian,
	title={Riemannian walk for incremental learning: Understanding forgetting and intransigence},
	author={Chaudhry, Arslan and Dokania, Puneet K and Ajanthan, Thalaiyasingam and Torr, Philip HS},
	booktitle={Proceedings of the European conference on computer vision (ECCV)},
	pages={532--547},
	year={2018}
}

@article{zhou2023deep,
	title={Deep class-incremental learning: A survey},
	author={Zhou, Da-Wei and Wang, Qi-Wei and Qi, Zhi-Hong and Ye, Han-Jia and Zhan, De-Chuan and Liu, Ziwei},
	journal={arXiv preprint arXiv:2302.03648},
	year={2023}
}

@article{li2017learning,
	title={Learning without forgetting},
	author={Li, Zhizhong and Hoiem, Derek},
	journal={IEEE transactions on pattern analysis and machine intelligence},
	volume={40},
	number={12},
	pages={2935--2947},
	year={2017},
	publisher={IEEE}
}

@inproceedings{hou2019learning,
	title={Learning a unified classifier incrementally via rebalancing},
	author={Hou, Saihui and Pan, Xinyu and Loy, Chen Change and Wang, Zilei and Lin, Dahua},
	booktitle={Proceedings of the IEEE/CVF conference on computer vision and pattern recognition},
	pages={831--839},
	year={2019}
}

@inproceedings{douillard2022dytox,
	title={Dytox: Transformers for continual learning with dynamic token expansion},
	author={Douillard, Arthur and Ram{\'e}, Alexandre and Couairon, Guillaume and Cord, Matthieu},
	booktitle={Proceedings of the IEEE/CVF Conference on Computer Vision and Pattern Recognition},
	pages={9285--9295},
	year={2022}
}

@inproceedings{zhao2020maintaining,
	title={Maintaining discrimination and fairness in class incremental learning},
	author={Zhao, Bowen and Xiao, Xi and Gan, Guojun and Zhang, Bin and Xia, Shu-Tao},
	booktitle={Proceedings of the IEEE/CVF conference on computer vision and pattern recognition},
	pages={13208--13217},
	year={2020}
}

@inproceedings{yan2021dynamically,
	title={Der: Dynamically expandable representation for class incremental learning},
	author={Yan, Shipeng and Xie, Jiangwei and He, Xuming},
	booktitle={Proceedings of the IEEE/CVF conference on computer vision and pattern recognition},
	pages={3014--3023},
	year={2021}
}

@inproceedings{wang2022foster,
	title={Foster: Feature boosting and compression for class-incremental learning},
	author={Wang, Fu-Yun and Zhou, Da-Wei and Ye, Han-Jia and Zhan, De-Chuan},
	booktitle={European conference on computer vision},
	pages={398--414},
	year={2022},
	organization={Springer}
}

@article{han2020automatically,
	title={Automatically discovering and learning new visual categories with ranking statistics},
	author={Han, Kai and Rebuffi, Sylvestre-Alvise and Ehrhardt, Sebastien and Vedaldi, Andrea and Zisserman, Andrew},
	journal={arXiv preprint arXiv:2002.05714},
	year={2020}
}

@article{han2021autonovel,
	title={Autonovel: Automatically discovering and learning novel visual categories},
	author={Han, Kai and Rebuffi, Sylvestre-Alvise and Ehrhardt, Sebastien and Vedaldi, Andrea and Zisserman, Andrew},
	journal={IEEE Transactions on Pattern Analysis and Machine Intelligence},
	volume={44},
	number={10},
	pages={6767--6781},
	year={2021},
	publisher={IEEE}
}

@article{zhao2021novel,
	title={Novel visual category discovery with dual ranking statistics and mutual knowledge distillation},
	author={Zhao, Bingchen and Han, Kai},
	journal={Advances in Neural Information Processing Systems},
	volume={34},
	pages={22982--22994},
	year={2021}
}

@inproceedings{vaze2022generalized,
	title={Generalized category discovery},
	author={Vaze, Sagar and Han, Kai and Vedaldi, Andrea and Zisserman, Andrew},
	booktitle={Proceedings of the IEEE/CVF Conference on Computer Vision and Pattern Recognition},
	pages={7492--7501},
	year={2022}
}

@article{fei2022xcon,
	title={Xcon: Learning with experts for fine-grained category discovery},
	author={Fei, Yixin and Zhao, Zhongkai and Yang, Siwei and Zhao, Bingchen},
	journal={arXiv preprint arXiv:2208.01898},
	year={2022}
}

@inproceedings{roy2022class,
	title={Class-incremental novel class discovery},
	author={Roy, Subhankar and Liu, Mingxuan and Zhong, Zhun and Sebe, Nicu and Ricci, Elisa},
	booktitle={European Conference on Computer Vision},
	pages={317--333},
	year={2022},
	organization={Springer}
}

@article{zhang2022grow,
	title={Grow and merge: A unified framework for continuous categories discovery},
	author={Zhang, Xinwei and Jiang, Jianwen and Feng, Yutong and Wu, Zhi-Fan and Zhao, Xibin and Wan, Hai and Tang, Mingqian and Jin, Rong and Gao, Yue},
	journal={Advances in Neural Information Processing Systems},
	volume={35},
	pages={27455--27468},
	year={2022}
}

@inproceedings{lee2013pseudo,
	title={Pseudo-label: The simple and efficient semi-supervised learning method for deep neural networks},
	author={Lee, Dong-Hyun and others},
	booktitle={Workshop on challenges in representation learning, ICML},
	volume={3},
	number={2},
	pages={896},
	year={2013},
	organization={Atlanta}
}

@article{rosenberg2005semi,
	title={Semi-supervised self-training of object detection models},
	author={Rosenberg, Chuck and Hebert, Martial and Schneiderman, Henry},
	year={2005},
	publisher={Carnegie Mellon University}
}

@article{zhang2021flexmatch,
	title={Flexmatch: Boosting semi-supervised learning with curriculum pseudo labeling},
	author={Zhang, Bowen and Wang, Yidong and Hou, Wenxin and Wu, Hao and Wang, Jindong and Okumura, Manabu and Shinozaki, Takahiro},
	journal={Advances in Neural Information Processing Systems},
	volume={34},
	pages={18408--18419},
	year={2021}
}

@article{khare2021unsupervised,
	title={Unsupervised class-incremental learning through confusion},
	author={Khare, Shivam and Cao, Kun and Rehg, James},
	journal={arXiv preprint arXiv:2104.04450},
	year={2021}
}

@inproceedings{he2021unsupervised,
	title={Unsupervised continual learning via pseudo labels},
	author={He, Jiangpeng and Zhu, Fengqing},
	booktitle={International Workshop on Continual Semi-Supervised Learning},
	pages={15--32},
	year={2021},
	organization={Springer}
}

@article{lloyd1982least,
	title={Least squares quantization in PCM},
	author={Lloyd, Stuart},
	journal={IEEE transactions on information theory},
	volume={28},
	number={2},
	pages={129--137},
	year={1982},
	publisher={IEEE}
}

@inproceedings{caron2018deep,
	title={Deep clustering for unsupervised learning of visual features},
	author={Caron, Mathilde and Bojanowski, Piotr and Joulin, Armand and Douze, Matthijs},
	booktitle={Proceedings of the European conference on computer vision (ECCV)},
	pages={132--149},
	year={2018}
}

@article{krizhevsky2009learning,
	title={Learning multiple layers of features from tiny images},
	author={Krizhevsky, Alex and Hinton, Geoffrey and others},
	year={2009},
	publisher={Toronto, ON, Canada}
}

@incollection{mccloskey1989catastrophic,
	title={Catastrophic interference in connectionist networks: The sequential learning problem},
	author={McCloskey, Michael and Cohen, Neal J},
	booktitle={Psychology of learning and motivation},
	volume={24},
	pages={109--165},
	year={1989},
	publisher={Elsevier}
}

@article{hinton2015distilling,
	title={Distilling the knowledge in a neural network},
	author={Hinton, Geoffrey and Vinyals, Oriol and Dean, Jeff},
	journal={arXiv preprint arXiv:1503.02531},
	year={2015}
}

@article{kuhn1955hungarian,
	title={The Hungarian method for the assignment problem},
	author={Kuhn, Harold W},
	journal={Naval research logistics quarterly},
	volume={2},
	number={1-2},
	pages={83--97},
	year={1955},
	publisher={Wiley Online Library}
}

@article{xu2005survey,
	title={Survey of clustering algorithms},
	author={Xu, Rui and Wunsch, Donald},
	journal={IEEE Transactions on neural networks},
	volume={16},
	number={3},
	pages={645--678},
	year={2005},
	publisher={Ieee}
}

@inproceedings{zhan2020online,
	title={Online deep clustering for unsupervised representation learning},
	author={Zhan, Xiaohang and Xie, Jiahao and Liu, Ziwei and Ong, Yew-Soon and Loy, Chen Change},
	booktitle={Proceedings of the IEEE/CVF conference on computer vision and pattern recognition},
	pages={6688--6697},
	year={2020}
}

@article{cui2018survey,
	title={A survey on network embedding},
	author={Cui, Peng and Wang, Xiao and Pei, Jian and Zhu, Wenwu},
	journal={IEEE transactions on knowledge and data engineering},
	volume={31},
	number={5},
	pages={833--852},
	year={2018},
	publisher={IEEE}
}

@article{cubuk2018autoaugment,
	title={Autoaugment: Learning augmentation policies from data},
	author={Cubuk, Ekin D and Zoph, Barret and Mane, Dandelion and Vasudevan, Vijay and Le, Quoc V},
	journal={arXiv preprint arXiv:1805.09501},
	year={2018}
}

@article{zhang2017mixup,
	title={mixup: Beyond empirical risk minimization},
	author={Zhang, Hongyi and Cisse, Moustapha and Dauphin, Yann N and Lopez-Paz, David},
	journal={arXiv preprint arXiv:1710.09412},
	year={2017}
}

@article{russakovsky2015imagenet,
	title={Imagenet large scale visual recognition challenge},
	author={Russakovsky, Olga and Deng, Jia and Su, Hao and Krause, Jonathan and Satheesh, Sanjeev and Ma, Sean and Huang, Zhiheng and Karpathy, Andrej and Khosla, Aditya and Bernstein, Michael and others},
	journal={International journal of computer vision},
	volume={115},
	pages={211--252},
	year={2015},
	publisher={Springer}
}

@inproceedings{kim2023proxy,
	title={Proxy anchor-based unsupervised learning for continuous generalized category discovery},
	author={Kim, Hyungmin and Suh, Sungho and Kim, Daehwan and Jeong, Daun and Cho, Hansang and Kim, Junmo},
	booktitle={Proceedings of the IEEE/CVF International Conference on Computer Vision},
	pages={16688--16697},
	year={2023}
}

@techreport{maji13fine-grained,
	title         = {Fine-Grained Visual Classification of Aircraft},
	author        = {S. Maji and J. Kannala and E. Rahtu
	and M. Blaschko and A. Vedaldi},
	year          = {2013},
	archivePrefix = {arXiv},
	eprint        = {1306.5151},
	primaryClass  = "cs-cv",
}

@inproceedings{krause20133d,
  title={3d object representations for fine-grained categorization},
  author={Krause, Jonathan and Stark, Michael and Deng, Jia and Fei-Fei, Li},
  booktitle={Proceedings of the IEEE international conference on computer vision workshops},
  pages={554--561},
  year={2013}
}

@inproceedings{parkhi2012cats,
  title={Cats and dogs},
  author={Parkhi, Omkar M and Vedaldi, Andrea and Zisserman, Andrew and Jawahar, CV},
  booktitle={2012 IEEE conference on computer vision and pattern recognition},
  pages={3498--3505},
  year={2012},
  organization={IEEE}
}

@article{paszke2019pytorch,
	title={Pytorch: An imperative style, high-performance deep learning library},
	author={Paszke, Adam and Gross, Sam and Massa, Francisco and Lerer, Adam and Bradbury, James and Chanan, Gregory and Killeen, Trevor and Lin, Zeming and Gimelshein, Natalia and Antiga, Luca and others},
	journal={Advances in neural information processing systems},
	volume={32},
	year={2019}
}

@inproceedings{he2016deep,
	title={Deep residual learning for image recognition},
	author={He, Kaiming and Zhang, Xiangyu and Ren, Shaoqing and Sun, Jian},
	booktitle={Proceedings of the IEEE conference on computer vision and pattern recognition},
	pages={770--778},
	year={2016}
}

@inproceedings{xu2020class,
	title={Class-weighted classification: Trade-offs and robust approaches},
	author={Xu, Ziyu and Dan, Chen and Khim, Justin and Ravikumar, Pradeep},
	booktitle={International conference on machine learning},
	pages={10544--10554},
	year={2020},
	organization={PMLR}
}

@inproceedings{welling2009herding,
	title={Herding dynamical weights to learn},
	author={Welling, Max},
	booktitle={Proceedings of the 26th annual international conference on machine learning},
	pages={1121--1128},
	year={2009}
}

@inproceedings{he2019rethinking,
	title={Rethinking imagenet pre-training},
	author={He, Kaiming and Girshick, Ross and Doll{\'a}r, Piotr},
	booktitle={Proceedings of the IEEE/CVF international conference on computer vision},
	pages={4918--4927},
	year={2019}
}

@inproceedings{belouadah2020scail,
	title={Scail: Classifier weights scaling for class incremental learning},
	author={Belouadah, Eden and Popescu, Adrian},
	booktitle={Proceedings of the IEEE/CVF winter conference on applications of computer vision},
	pages={1266--1275},
	year={2020}
}

@inproceedings{NEURIPS2021_77ee3bc5,
	author = {Zhu, Fei and Cheng, Zhen and Zhang, Xu-yao and Liu, Cheng-lin},
	booktitle = {Advances in Neural Information Processing Systems},
	editor = {M. Ranzato and A. Beygelzimer and Y. Dauphin and P.S. Liang and J. Wortman Vaughan},
	pages = {14306--14318},
	publisher = {Curran Associates, Inc.},
	title = {Class-Incremental Learning via Dual Augmentation},
	url = {https://proceedings.neurips.cc/paper_files/paper/2021/file/77ee3bc58ce560b86c2b59363281e914-Paper.pdf},
	volume = {34},
	year = {2021}
}

@article{karpusha2020calibrated,
	title={Calibrated neighborhood aware confidence measure for deep metric learning},
	author={Karpusha, Maryna and Yun, Sunghee and Fehervari, Istvan},
	journal={arXiv preprint arXiv:2006.04935},
	year={2020}
}

@article{sikar2024accept,
	title={When to Accept Automated Predictions and When to Defer to Human Judgment?},
	author={Sikar, Daniel and Garcez, Artur and Weyde, Tillman and Bloomfield, Robin and Peeroo, Kaleem},
	journal={arXiv preprint arXiv:2407.07821},
	year={2024}
}

@inproceedings{wu2024ft,
	title={FT K-Means: A High-Performance K-Means on GPU with Fault Tolerance},
	author={Wu, Shixun and Ding, Yitong and Zhai, Yujia and Liu, Jinyang and Huang, Jiajun and Jian, Zizhe and Dai, Huangliang and Di, Sheng and Wong, Bryan M and Chen, Zizhong and others},
	booktitle={2024 IEEE International Conference on Cluster Computing (CLUSTER)},
	pages={322--334},
	year={2024},
	organization={IEEE}
}

@article{li2023large,
	title={Large scale K-means clustering using GPUs},
	author={Li, Mi and Frank, Eibe and Pfahringer, Bernhard},
	journal={Data Mining and Knowledge Discovery},
	volume={37},
	number={1},
	pages={67--109},
	year={2023},
	publisher={Springer}
}

@article{pedregosa2011scikit,
	title={Scikit-learn: Machine learning in Python},
	author={Pedregosa, Fabian and Varoquaux, Ga{\"e}l and Gramfort, Alexandre and Michel, Vincent and Thirion, Bertrand and Grisel, Olivier and Blondel, Mathieu and Prettenhofer, Peter and Weiss, Ron and Dubourg, Vincent and others},
	journal={the Journal of machine Learning research},
	volume={12},
	pages={2825--2830},
	year={2011},
	publisher={JMLR. org}
}

@article{giffon2021quick,
	title={QuicK-means: accelerating inference for K-means by learning fast transforms},
	author={Giffon, Luc and Emiya, Valentin and Kadri, Hachem and Ralaivola, Liva},
	journal={Machine Learning},
	volume={110},
	number={5},
	pages={881--905},
	year={2021},
	publisher={Springer}
}

@article{von2007tutorial,
  title={A tutorial on spectral clustering},
  author={Von Luxburg, Ulrike},
  journal={Statistics and computing},
  volume={17},
  pages={395--416},
  year={2007},
  publisher={Springer}
}

@article{mullner2011modern,
  title={Modern hierarchical, agglomerative clustering algorithms},
  author={M{\"u}llner, Daniel},
  journal={arXiv preprint arXiv:1109.2378},
  year={2011}
}

@inproceedings{pu2023dynamic,
  title={Dynamic conceptional contrastive learning for generalized category discovery},
  author={Pu, Nan and Zhong, Zhun and Sebe, Nicu},
  booktitle={Proceedings of the IEEE/CVF conference on computer vision and pattern recognition},
  pages={7579--7588},
  year={2023}
}

@inproceedings{choi2024contrastive,
  title={Contrastive mean-shift learning for generalized category discovery},
  author={Choi, Sua and Kang, Dahyun and Cho, Minsu},
  booktitle={Proceedings of the IEEE/CVF Conference on Computer Vision and Pattern Recognition},
  pages={23094--23104},
  year={2024}
}
}
\newpage
\appendix
\onecolumn
\section{Overview}

\begin{figure*}[t]
	\centering
	\includegraphics[width=0.9\textwidth]{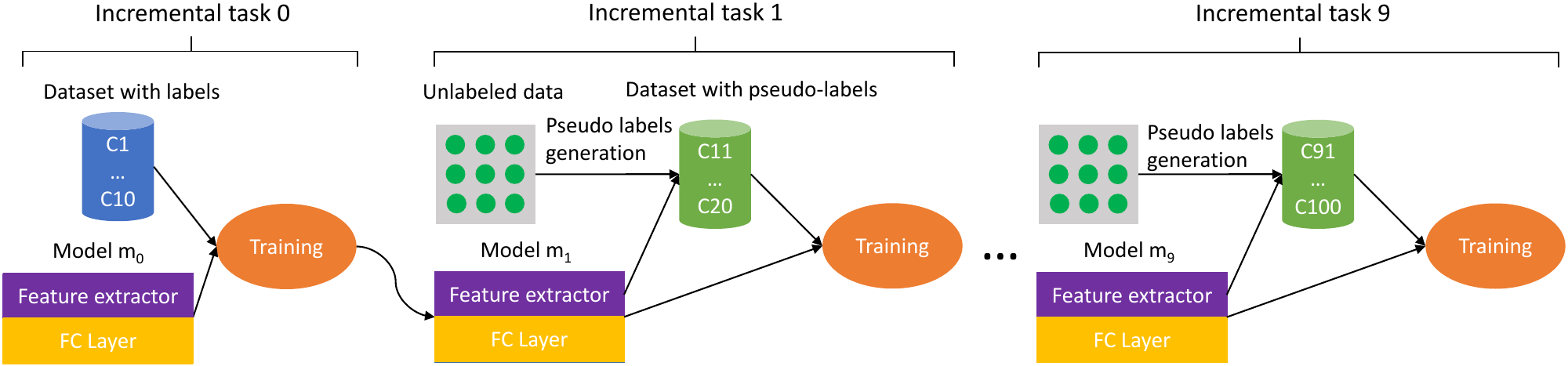}
	\caption{An overview of ICPL.}
	\label{overview}
\end{figure*}

Figure \ref{overview} provides an overview of ICPL. Incremental task 0 is the same as in supervised incremental learning. Then, we use the feature extractor and the unlabeled data to generate pseudo-labels with the methodology shown in Figure \ref{pseudo}. These pseudo-labels can be incorporated into popular CIL methods, and this process is repeated until there is no more data.

\section{Ablation study}
\label{app:ablation}
\subsection{Hyperparameter choice}

We conducted extensive experiments to examine the impact of hyperparameters on the ICPL method. Firstly, we evaluated the impact of step size on the performance of our ICPL method. Table \ref{step} presents these results, showing that a step size of 10 yields the best performance, while larger step sizes result in performance degradation. Notably, the scenario with no step size, where pseudo-labels are never recomputed during incremental training, produced the lowest performance. This underscores the importance of periodically updating pseudo-labels during training to improve both their quality and quantity.

Next, we conducted an ablation study to assess the impact of AutoAugment \cite{cubuk2018autoaugment} data augmentation on the performance of our method. Table \ref{autoaugment} summarizes the results, showing that nearly all configurations perform better with this data augmentation. This suggests that AutoAugment may enhance the model's discrimination capability, thereby improving the quality of the pseudo-labels.

\begin{table}[b]
	\centering
    \small
	\begin{tabular}{@{}cccccccccc@{}}
		\toprule
		\multirow{2}{*}{Method} & \multirow{2}{*}{Step size $\tau$} & \multicolumn{2}{c}{Base0 Inc5} & \multicolumn{2}{c}{Base0 Inc10} & \multicolumn{2}{c}{Base0 Inc20} & \multicolumn{2}{c}{Base50 Inc10} \\ \cmidrule(l){3-4} \cmidrule(l){5-6} \cmidrule(l){7-8} \cmidrule(l){9-10}
		& & $\Theta_N$           & $\tilde{\Theta}$          & $\Theta_N$           & $\tilde{\Theta}$ &  $\Theta_N$           & $\tilde{\Theta}$ & $\Theta_N$           & $\tilde{\Theta}$                         \\ \midrule \midrule
        \multirow{4}{*}{Replay} & $\varnothing$       & 26.30 & 42.53    & 23.95              & 39.57 & 29.38 & \textbf{50.30}           & \textbf{37.68}          & \textbf{50.65}   \\
		& 10        & \textbf{30.02} & 40.75        &  \textbf{31.57}           & \textbf{48.40}        & \textbf{30.13} & 46.76 & 30.01             & 40.27                                        \\
		& 20    & 27.92 & 46.25               & 23.85         & 41.65   & 24.13 & 44.30     & 32.43          & 47.49              \\
		& 30     & 29.67 &\textbf{48.61}             & 25.27             & 42.12   & 22.36 & 43.68      &  31.76            & 46.79 \\ \midrule
        \multirow{4}{*}{iCaRL} & $\varnothing$       & 27.21 & 42.75    & 25.91             & 42.06 & 27.33 & 46.70           & \textbf{37.80}              & \textbf{52.95}  \\
		& 10        & \textbf{30.79} & 41.16        & \textbf{32.03}            & \textbf{49.00}         & \textbf{33.32} & \textbf{50.86} &   37.30            & 48.15                                         \\
		& 20    & 28.44 & \textbf{46.50}               & 25.06               & 43.20   & 26.03 & 46.67     & 36.56           & 51.47               \\
		& 30     & 30.06 & 46.01             & 25.45            & 43.62   & 26.52 & 47.02      & 34.11           & 50.92 \\ \midrule
        \multirow{4}{*}{WA} & $\varnothing$       & 32.50 & 44.92    & 29.94              & 44.72 & 34.96 & 51.08           & 45.38             & 57.51   \\
		& 10        & 33.78 & \textbf{51.87}        & \textbf{33.50}            & \textbf{48.65}         & \textbf{36.78} & \textbf{53.22} &   \textbf{45.86}            & 57.74                                         \\
		& 20    & \textbf{37.03} & 50.18               & 31.66              & 47.30     & 35.73 & 51.52     &   44.62            & \textbf{58.01}               \\
		& 30     & 34.66 & 48.18             & 32.03              & 47.33    & 35.33 & 51.69      &   44.75            & 57.12 \\ \midrule
        \multirow{4}{*}{FOSTER} & $\varnothing$       & 30.53 & 42.39    & 28.88             & 44.39 & 39.57 & 54.38           & 53.35             & 65.01   \\
		& 10        & \textbf{39.23} & \textbf{53.24}        & \textbf{35.57}            & \textbf{50.00}         & \textbf{40.90} & \textbf{55.64} &  \textbf{55.13}            & \textbf{65.68}                                         \\
		& 20    & 38.39 & 51.02               & 34.89             & 49.04    & 39.82 & 54.85     & 53.46             &  65.02              \\
		& 30     & 35.93 & 49.78             & 33.76            & 48.24   & 39.39 & 54.02      & 53.87        & 65.45 \\\bottomrule                                                                             
	\end{tabular}
    \caption{Ablation study for the choice of step size on CIFAR100. The model used is WA + ICPL for each configurations.}
	\label{step}
\end{table}

Finally, Table \ref{confidence} highlights how different confidence thresholds affect the results across various configurations. On average, the default threshold of 0.85 provides the best overall performance, while lowering this threshold slightly reduces performance. A threshold of 0.95 yields the best results for the Base50 Inc10 configuration but performs poorly for Base0 Inc10. In the latter case, with few base classes, the model struggles to form clusters that meet the high threshold requirement. This study demonstrates that the confidence threshold significantly influences performance, as it balances the tradeoff between the quality and quantity of pseudo-labels.

\begin{table*}[t]
	\centering
    \footnotesize
	\begin{tabular}{@{}ccccccccccc@{}}
		\toprule
		\multirow{2}{*}{Method} & \multirow{2}{*}{ICPL} & \multirow{2}{*}{AutoAugment} & \multicolumn{2}{c}{Base0 Inc5} & \multicolumn{2}{c}{Base0 Inc10} & \multicolumn{2}{c}{Base0 Inc20} & \multicolumn{2}{c}{Base50 Inc10} \\ \cmidrule(l){4-5} \cmidrule(l){6-7} \cmidrule(l){8-9} \cmidrule(l){10-11}
		& & & $\Theta_N$           & $\tilde{\Theta}$          & $\Theta_N$           & $\tilde{\Theta}$           & $\Theta_N$           & $\tilde{\Theta}$           & $\Theta_N$            & $\tilde{\Theta}$                    \\ \midrule \midrule
		\multirow{2}{*}{Replay} & \checkmark & \checkmark           & \textbf{30.02}              & 40.75             & \textbf{31.57}              & \textbf{48.40}              & \textbf{30.13}              & \textbf{46.76}              & 30.01               & 40.27                          \\
		          & \checkmark &        & 29.75              & \textbf{43.42}             & 27.8              & 44.48              & 21.68              & 38.32              & \textbf{35.43}               & \textbf{47.87}                           \\ \midrule
		\multirow{2}{*}{iCaRL} & \checkmark & \checkmark          & \textbf{30.79}             &  41.16            & \textbf{32.03}              & \textbf{49.00}              & \textbf{33.32}              & \textbf{50.86}              & \textbf{37.3}               & 48.15                           \\
		        & \checkmark &                       & 27.35              & \textbf{43.22}             & 30.22              & 45.90              & 30.75              & 47.63              & 36.63               & \textbf{50.75}                           \\ \midrule
		\multirow{2}{*}{WA} & \checkmark & \checkmark              & \textbf{33.78}           & \textbf{51.87}            & \textbf{33.5}            & \textbf{48.65}            & \textbf{36.78}           & \textbf{53.22}            & \textbf{45.86}            & \textbf{57.74}                      \\
		      & \checkmark &                          & 25.98              & 39.38             & 27.23              & 41.66              & 30.99              & 48.22              & 43.15               & 55.29                           \\ \midrule
		\multirow{2}{*}{FOSTER} & \checkmark & \checkmark                 & \textbf{39.23}           & \textbf{53.24}            & \textbf{35.57}           & \textbf{50.00}            & \textbf{40.9}            & \textbf{55.64}           & \textbf{55.13}             & \textbf{65.68}   \\
		   & \checkmark &                              & 29.39              & 43.26             & 29.12              & 43.28              & 33.59              & 49.70              & 50.47               & 62.31                          \\ \bottomrule
	\end{tabular}
    \caption{Comparison of average accuracy $\tilde{\Theta}$ and final accuracy $\Theta_N$ on CIFAR100 with ICPL, with and without AutoAugment data augmentation.}
	\label{autoaugment}
\end{table*}

\begin{table}[]
	\centering
    \scriptsize
	\begin{tabular}{@{}cccccccccccccccccc@{}}
		\toprule
		\multirow{2}{*}{Confidence threshold $\alpha$} & \multicolumn{3}{c}{Base0 Inc5} & \multicolumn{3}{c}{Base0 Inc10} & \multicolumn{3}{c}{Base0 Inc20} &\multicolumn{3}{c}{Base50 Inc10} \\ \cmidrule(l){2-4} \cmidrule(l){5-7} \cmidrule(l){8-10} \cmidrule(l){11-13}
		& $\Theta_N$           & $\tilde{\Theta}$          & \% & $\Theta_N$           & $\tilde{\Theta}$ & \% & $\Theta_N$           & $\tilde{\Theta}$ & \% & $\Theta_N$           & $\tilde{\Theta}$ & \%                            \\ \midrule \midrule
		0.65          & 36.00 & 52.08 & 19  & 30.53              & 47.35  & 31          & 37.15 & 53.46 & 38 & 44.52              & 57.73  &   16                                   \\
		0.75      & \textbf{36.81} & \textbf{52.52} &  36       & 32.97              & 47.52  & 50  & 36.37 & 52.95 & 55       & 45.39              & 57.61   & 30                                     \\
		0.85            & 33.78 & 51.87 & 60        & \textbf{33.50}            & \textbf{48.65}       & 42 & \textbf{37.78} & \textbf{53.52} &  73  &   45.86            & 57.74  & 30            \\
		0.95     & 36.71 & 50.77 & 78              & 29.35              & 43.75  & 86          & 36.28 & 52.58 & 89 &  \textbf{47.14}            & \textbf{57.89} & 59 \\ \bottomrule                                       
	\end{tabular}
    \caption{Impact of varying confidence thresholds on average accuracy $\tilde{\Theta}$, final accuracy $\Theta_N$ and percentage of skipped samples $\%$ for the CIFAR100 dataset. The model used is WA + ICPL for each configurations.}
	\label{confidence}
\end{table}

\subsection{Impact of strong data augmentation and class-weight balancing on various incremental learning method}
\label{app:subsec:data}

During our experiments, we discovered that class-weight balancing and aggressive data augmentation techniques like MixUp can improve performance for some CIL methods, but not all. Some incremental learning methods are incompatible with these techniques. Table \ref{strong} presents the results for various CIL methods with and without class-weight balancing \cite{xu2020class} and MixUp \cite{zhang2017mixup}. From these results, we can observe that Replay \cite{zhou2023deep} and iCaRL \cite{rebuffi2017icarl} methods generally benefit from these techniques, while WA \cite{zhao2020maintaining} and FOSTER \cite{wang2022foster} exhibit a decline in performance.

Class-weight balancing and MixUp are proposed solutions \cite{belouadah2020scail, NEURIPS2021_77ee3bc5} to address catastrophic forgetting. However, CIL methods with higher complexity, such as WA and FOSTER, do not benefit from these techniques because they already partially solve the issues that class-weight balancing and MixUp aim to address. WA assumes that the model will be biased toward new classes and adjusts the classifier weights accordingly, so when class-weight balancing is applied, this problem no longer arises. FOSTER is a dynamic method that can allocate parts of its model to learn new classes, eliminating the need for MixUp to enhance the retention of knowledge of old classes, as specific backbones manage this task.

Conversely, simpler methods like Replay and iCaRL benefit from class-weight balancing and MixUp because these techniques address problems that are not fully resolved by data replay and knowledge distillation alone.

\begin{table*}[]
	\centering
    \scriptsize
	\begin{tabular}{@{}cccccccccccc@{}}
		\toprule
		\multirow{2}{*}{Method} & \multirow{2}{*}{ICPL} & \multirow{2}{*}{Class Weight} & \multirow{2}{*}{MixUp} & \multicolumn{2}{c}{Base0 Inc5} & \multicolumn{2}{c}{Base0 Inc10} & \multicolumn{2}{c}{Base0 Inc20} & \multicolumn{2}{c}{Base50 Inc10} \\ \cmidrule(l){5-6} \cmidrule(l){7-8} \cmidrule(l){9-10} \cmidrule(l){11-12}
		& & & & $\Theta_N$           & $\tilde{\Theta}$          & $\Theta_N$           & $\tilde{\Theta}$           & $\Theta_N$           & $\tilde{\Theta}$           & $\Theta_N$            & $\tilde{\Theta}$                    \\ \midrule \midrule
		\multirow{3}{*}{Replay} & \checkmark &          &         & \textbf{30.54}              & \textbf{48.51}             & 24.6              & 43.26              & 24.49              & 43.70              & 32.42               & 47.20                          \\
		& \checkmark & \checkmark &      & 27.25              & 44.13             & 27.16              & 42.89              & 25.62              & 44.00              & \textbf{35.56}               & \textbf{48.55}                          \\
		& \checkmark & \checkmark & \checkmark        & 30.02              & 40.75             & \textbf{31.57}              & \textbf{48.40}              & \textbf{30.13}              & \textbf{46.76}              & 30.01               & 40.27                          \\ \midrule
		\multirow{3}{*}{iCaRL} & \checkmark &          &                 & 28.37              & 46.06             & 25.3              & 43.19              & 25.11              & 46.42              & 35.75               & \textbf{51.47}                          \\
		& \checkmark & \checkmark &         & \textbf{31.65}              & \textbf{46.61}             & 27.63              & 44.48              & 28.27              & 48.23              & 35.46               & 49.19                          \\
		& \checkmark & \checkmark & \checkmark        & 30.79             &  41.16            & \textbf{32.03}              & \textbf{49.00}              & \textbf{33.32}              & \textbf{50.86}              & \textbf{37.3}               & 48.15                          \\ \midrule
		\multirow{3}{*}{WA} & \checkmark &          &                    & 33.78           & 51.87            & \textbf{33.5}            & \textbf{48.65}            & \textbf{36.78}           & \textbf{53.22}            & \textbf{45.86}            & \textbf{57.74}                          \\
		& \checkmark & \checkmark &         & 33.46              & 48.92             & 32.03              & 47.79              & 31.3              & 50.28              & 39.04               & 52.76                          \\
		& \checkmark & \checkmark & \checkmark        & \textbf{36.75}              & \textbf{54.96}             & 30.15              & 46.60              & 32.0              & 49.50              & 37.57               & 49.79                          \\ \midrule
		\multirow{3}{*}{FOSTER} & \checkmark &          &                   & \textbf{39.23}           & \textbf{53.24}            & 35.57           & 50.00            & \textbf{40.9}            & \textbf{55.64}           & \textbf{55.13}             & 65.68                          \\
		& \checkmark & \checkmark &             & 37.55              & 51.95             & \textbf{35.58}              & \textbf{51.66}              & 38.65              & 55.00              & 55.12              & \textbf{66.39}                          \\
		& \checkmark & \checkmark & \checkmark            & 37.02              & 52.40             & 32.96              & 49.10              & 36.19              & 52.68              & 50.55               & 62.36                          \\ \bottomrule
	\end{tabular}
    \caption{Analysis of the impact of class-weight balancing and MixUp methods for various CIL methods}
	\label{strong}
\end{table*}

\section{GFLOPs analysis}
\label{app:subsec:gflop}

In this paper, we studied the computational cost of our ICPL method, showing it reduces complexity compared to the supervised approach by excluding low-confidence pseudo-labels. Here, we analyze the cost of a single incremental step using the Base10 Inc10 setup on the CIFAR100 dataset, comparing supervised and unsupervised approaches. Our results provide an approximate GFLOPs value, excluding the computational cost of memory management, loss computation, and the optimizer. However, these operations are negligible compared to the total training cost, making our approximation reasonable. 

In the supervised case, the number of samples, \(n_{\text{supervised}}\), is 7000, including 5000 images from the 10 new classes and a 2000-image rehearsal dataset. In the unsupervised case with ICPL, \(n_{\text{unsupervised}}\) is approximated as 5000, combining the rehearsal dataset and 3000 high-confidence pseudo-labeled images. Since pseudo-labels are recomputed every 10 epochs, the 3000 number is an approximation used for this analysis. The number of epochs, $\text{epochs}$, is set to 170 and the step size, $\tau$, is set to 10.

As proposed by \cite{wu2024ft, giffon2021quick}, the computational cost of KMeans operations can be calculated using Equation \ref{eq:1}:

\begin{equation} \label{eq:1}
	\text{KMeans}_{\text{GFLOPs}} = I \times n \times d \times k
\end{equation}

where \( I \) represents the number of iterations, \( n \) the number of data points, \( d \) the dimensionality of the embeddings, and \( k \) the number of clusters. In our case study with the Base10 Inc10 setup, maximum number of iterations of the KMeans algorithm is fixed at 100, though it typically falls between 40 and 60 (we use \(I\! =\! 50\) in this example). Here, \( n\! =\! 5000 \) (corresponding to 10 classes), \( d\! =\! 64 \), and \( k\! =\! 10 \). We therefore obtain $\text{KMeans}_{\text{GFLOPs}}\! =\! 0.16\, \text{GFLOPs}$.

For our ResNet32 model, the GFLOPs for a forward + backward (Training) step and an forward (Inference) step per image are $\text{Training}_{\text{GFLOPs}}\! =\! 0.41$ and $\text{Inference}_{\text{GFLOPs}}\! =\! 0.14$, respectively. 
Using these values, the GFLOPs required for pseudo-label generation can be calculated as shown in Equation \ref{eq:2}:

\begin{equation} \label{eq:2}
	\begin{split}
	\text{Pseudo-Labels}_{\text{GFLOPs}} & = \text{Inference}_{\text{GFLOPs}} \times n +  \text{KMeans}_{\text{GFLOPs}} \\
	& = 688.66\, \text{GFLOPs}
\end{split}
\end{equation}

For one incremental step in the supervised case, the total computational cost is derived using Equation \ref{eq:3}:

\begin{equation} \label{eq:3}
	\begin{split}
	\text{Supervised}_{\text{GFLOPs}} & = (n_{\text{supervised}} \times \text{epochs}) \times \text{Training}_{\text{GFLOPs}} \\ & = 487900\, \text{GFLOPs}
\end{split}
\end{equation}

In comparison, for the unsupervised case, the total computational cost is calculated using Equation \ref{eq:4}:

\begin{equation}  \label{eq:4}
	\begin{split}
	\text{Unsupervised}_{\text{GFLOPs}}  & = (n_{\text{unsupervised}} \times \text{epochs}) \times \text{Training}_{\text{GFLOPs}} \\ 
	& + (1 + \left\lfloor \frac{\text{epochs}}{\tau} \right\rfloor) \times \text{Pseudo-Labels}_{\text{GFLOPs}} \\
	& = 360207\, \text{GFLOPs}
\end{split}
\end{equation}

This demonstrates that our ICPL method is computationally efficient, achieving a 26.17\% reduction in computational cost compared to the supervised approach. In addition, the pseudo-labels generation incurs a cost of only 11707 GFLOPs, which is relatively low compared to the total training cost of 360173 GFLOPs, accounting for just 3.25\% of the total.

\section{NMI and ARI evaluation}

Normalized Mutual Information (NMI) and Adjusted Rand Index (ARI) are two metrics used to evaluate the quality of agreement between two clusterings, making them particularly useful in our unsupervised approach. 

The NMI value between two clusters, U and V, is defined as follows

\[ NMI(U, V)\! =\! \frac{I(U, V)}{\sqrt{H(U) \cdot H(V)}} \]

where $I(U, V)$ represents the mutual information between U and V and $H(U)$ represents the entropy of cluster U. The NMI value ranges from 0 to 1, where a value of 1 represents a perfect correlation between the two partitions, and a value of 0 indicates no mutual information.

The ARI between two clusters, U and V, is defined as follows:

\[ ARI\! =\! \frac{ RI - E[RI] }{ \max(RI) - E[RI] } \]

where $RI$ is the Rand Index, $E[RI]$ represents the expected value of the Rand Index when clusterings are assigned randomly and $max(RI)$ denotes the highest possible value that the Rand Index can achieve.

ARI is designed to be close to 0.0 for random labelings, regardless of the number of clusters or samples, and it reaches exactly 1.0 when the clusterings are identical (up to permutation). Additionally, the ARI is bounded below by -0.5 for particularly discordant clusterings.

\begin{figure*}[]
	\centering
	\includegraphics[width=0.85\textwidth]{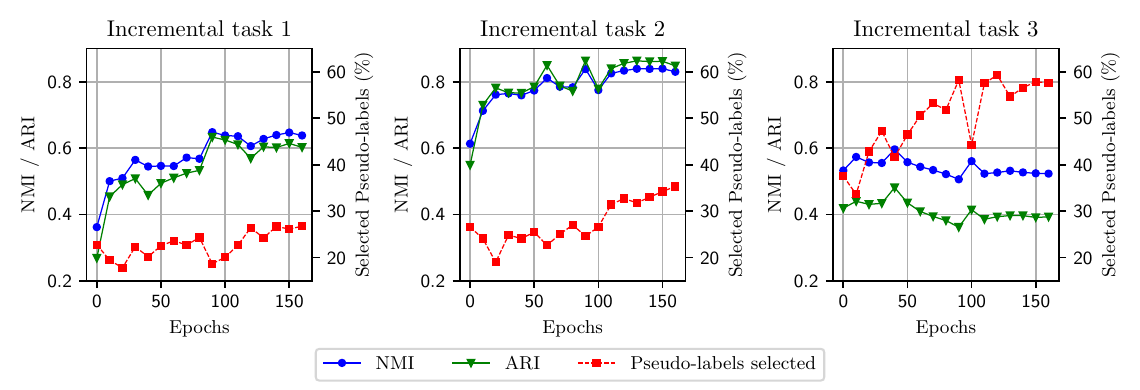}
	\caption{Graph depicting the evolution of NMI and ARI between selected pseudo-labels and ground-truth labels over epochs, for various incremental steps. The percentage of selected pseudo-labels is displayed in red. The method used is WA + ICPL on a Base10 Inc10 setup on CIFAR100}
	\label{nmi}
\end{figure*}

Figure \ref{nmi} illustrates how the NMI and ARI between selected pseudo-labels and ground-truth labels evolve over epochs across various incremental steps, along with the percentage of selected pseudo-labels. Overall, both the quality (as indicated by NMI and ARI) and the quantity (as represented by the percentage of selected pseudo-labels) improve with additional epochs, highlighting the benefit of recomputing pseudo-labels throughout the incremental task.

\end{document}